# Efficient Selection of Disambiguating Actions for Stereo Vision

**Monika Schaeffer and Ronald Parr**
Duke University
Department of Computer Science
{monika,parr}@cs.duke.edu

## Abstract

In many domains that involve the use of sensors, such as robotics or sensor networks, there are opportunities to use some form of active sensing to disambiguate data from noisy or unreliable sensors. These disambiguating actions typically take time and expend energy. One way to choose the next disambiguating action is to select the action with the greatest expected entropy reduction, or information gain. In this work, we consider active sensing in aid of stereo vision for robotics. Stereo vision is a powerful sensing technique for mobile robots, but it can fail in scenes that lack strong texture. In such cases, a structured light source, such as vertical laser line, can be used for disambiguation. By treating the stereo matching problem as a specially structured HMM-like graphical model, we demonstrate that for a scan line with $n$ columns and maximum stereo disparity $d$, the entropy minimizing aim point for the laser can be selected in $O(nd)$ time - cost no greater than the stereo algorithm itself. A typical HMM formulation would suggest at least $O(nd^2)$ time for the entropy calculation alone.

## 1 Introduction

In many domains that involve the use of sensors, such as robotics or sensor networks, there are opportunities to use some form of active sensing to disambiguate data from noisy or unreliable sensors. These disambiguating actions typically take time and expend energy, so some care must be exercised to choose these actions wisely.

In the field of robotics, laser range finders have been the sensor of choice in recent years due to their great accuracy. However, there are many disadvantages to laser range finders. Even the two-dimensional models are expensive and bulky, with calibrated moving parts that consume a lot of power. They are also far from stealthy, an important consideration for some applications. Three-dimensional laser range finders share the shortcomings of their two-dimensional counterparts at ten to twenty times the cost.

Recent advances in camera technology have made cameras an inexpensive and versatile sensor for many applications. When used in a stereo pair, they have the potential to replace laser range finders as accurate depth sensing devices since recent advances in the pixel density of cameras can, in principle, permit more accurate depth estimates over larger distances.

While standard stereo algorithms are known to perform well on some highly textured benchmark images [7], they have trouble in scenes with little texture, such as the long blank walls often encountered in indoor robotics applications. One way to address this problem is to introduce texture into the scene through the use of a structured light source. If size, time, and stealth are not considerations, and the target is neither far away nor brightly illuminated, a sequence of computer generated patterns from a projector can be used for disambiguation [8]. However, in many applications, such a blanket approach is not practical.

In this paper, we consider a more concentrated light source, specifically a laser line projector, mounted on a pan-tilt head. A laser line projector can cover an entire vertical stripe through a scene with fairly bright light, yet it is compact and inexpensive. Moving the laser and projecting a laser line for each of the thousands of columns in a high resolution image would be rather time consuming. Since information propagates horizontally though the stereo algorithm, precise control of the pan-tilt head is not needed – a laser aim improves accuracy in many of the pixels adjacent to the area struck by the laser. In addition, we integrate the laser into a stereo vision algorithm for a cycle that first estimates stereo disparity, then selects and optimal query point for the laser, then repeats – hopefully as few times as needed to get good estimates of depth.

The problem of selecting the best information gathering action is, in general, a quite difficult form of metacomputa-tion. A poor technique for deciding where to aim the laser could easily spend more time deliberating than it would take to sweep the laser through every column in the image. Here we consider the aim point for the laser with the greatest expected entropy reduction, or information gain, in the space of stereo matchings. This is a myopic procedure, but it does quite well in practice and, importantly, it can be computed very efficiently. By converting an existing dynamic programming stereo vision approach to an equivalent, highly structured HMM problem, we can then apply insights from HMM information gain computations to the problem, allowing us to calculate the expected entropy reduction of each potential laser line very efficiently. We demonstrate that for a scan line with $n$ columns and maximum stereo disparity $d$, the entropy minimizing aim point for the laser can be selected in $O(nd)$ time - cost no greater than the stereo algorithm itself. In contrast, a typical HMM formulation would suggest at least $O(nd^2)$ time for the entropy calculation alone.

We present results of this approach on synthetic benchmark problems and on a set of images collected by a prototype device that implements these ideas.

## 2 Augmenting Stereo Vision with Lasers

A typical stereo algorithm assumes that its input images are captured by two capture media, e.g. solid state sensors, that are in the same plane, have identical (ideal) lenses, and are aligned so that corresponding rows of pixels in each image are on the same line. Even if this assumption is not entirely true (as it rarely is), it is assumed that the images are rectified in software to approximate this ideal. Pixel-to-pixel stereo algorithms work by finding, for each pixel in one frame, a corresponding pixel in the other frame (or declaring the pixel to be occluded). The assumption of a calibrated camera system limits the scope of this search to pixels along the corresponding scan line in the adjacent frame, thereby permitting relatively efficient stereo algorithms. When a correct match is found, the depth can recovered straightforwardly from the difference in horizontal offsets of the pixels (the *disparity*) and the geometry of the camera system.

Bobick and Intille [3] present stereo vision as a shortest path problem through a data structure called the *disparity space image* (*DSI*), which, for a horizontal scanline across a stereo image pair, captures information about all potential matches between left and right image pixels. (In their simplest form, stereo algorithms assume independence between horizontal scan lines. Elaborations are, of course, both possible and frequent.) The *DSI* is an $n \times d$ matrix of cells, each of which may take one of three possible values. There are no more than 7 transitions in and out of each cell and no cycles. Edge costs between cells correspond to either a match quality score based upon the luminance

i: location of the pixel in the right image
$j$: disparity value
$M$: A matched pixel
$L$: A left image occlusion with penalty $D_l$
$R$: A right image occlusion with penalty $D_r$
*score*$(i, j)$: pixel value difference between images of difference between the pixels that are matched, or occlusion penalty score for failing to match a pixel. Thus, the lowest cost path through the graph corresponds to the best stereo matching. The main advantage of the DSI view of the stereo problem is that the sparseness and regularity of the DSI structure permit a dynamic programming solution to the shortest path problem in a single scan line in $O(nd)$ time. This view also implicitly encodes two constraints on the space of possible matchings, the *uniqueness* constraint and the *ordering* constraint. The uniqueness constraint permits each pixel to match at most one other pixel. The ordering constraint requires that the indices of matches along any row of the image are non-decreasing (increasing when combined with the uniqueness constraint). The consequences of these assumptions are discussed in more detail in Section 2.4.

| from node | to node | cost |
|---|---|---|
| (i,j,M) | (i,j-1,L) | $D_l$ |
| (i,j,M) | (i+1,j,M) | *score*(i+1,j) |
| (i,j,M) | (i+1,j+1,R) | $D_r$ |
| (i,j,L) | (i,j-1,R) | $D_l$ |
| (i,j,L) | (i+1,j,M) | *score*(i+1,j) |
| (i,j,R) | (i+1,j,M) | *score*(i+1,j) |
| (i,j,R) | (i+1,j+1,R) | $D_r$ |

Figure 1: Node transition costs when viewing the DSI as a graphical model.

### 2.1 The Disparity Space Image

We view the *DSI* as a graphical model with $n \times d \times 3$ nodes and the arc costs shown in Figure 1. The graphical structure is shown in Figure 2. The path costs in the original DSI induce a measure on the space of matchings. We therefore convert the DSI costs to unnormalized potentials in a graphical model by exponentiating negated scores. (In this view, a score that is a squared luminance difference corresponds to a Gaussian sensor noise model.) We can now view the DSI as defining an HMM-like chain graphical model where the individual states have internal structure. The set of nodes $(i, *, *)$ collectively define the distribution over the disparity values of pixel $i$ in the image and correspond to a single HMM state. The arcs between $(i, *, *)$ and $(i + 1, *, *)$ define the (non-stationary) transition probabilities between adjacent states.

The benefit of this representation over a standard HMM formulation is that we can use a modified version of the forward-backward algorithm that exploits the internal state

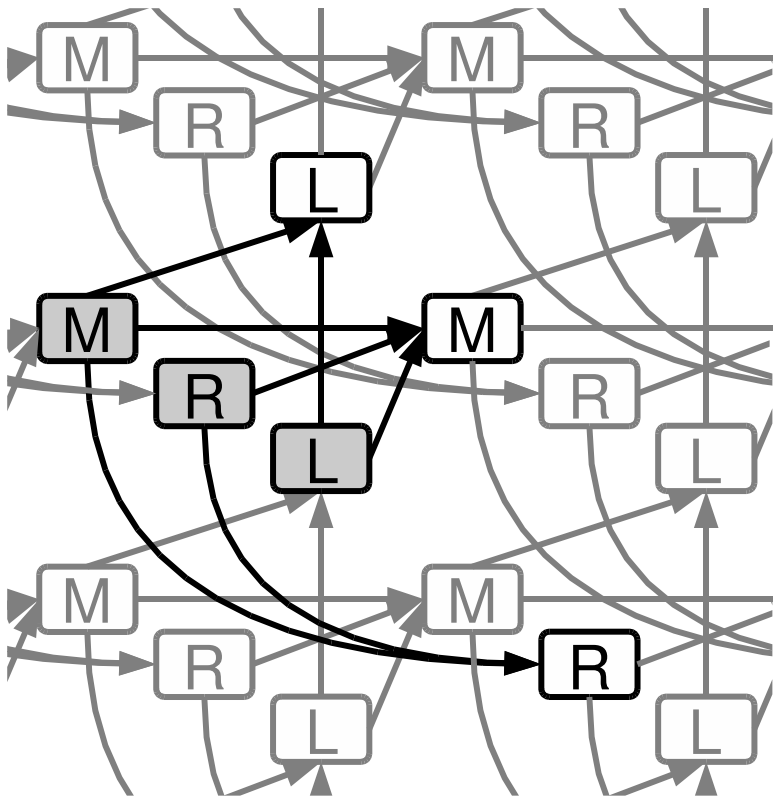

Figure 2: A collection of nodes internal to the DSI, with the seven transitions out of a triplet of nodes highlighted. This pattern repeats to the ends of the DSI. Within a column, M and R nodes are jointly exhaustive and mutually exclusive. The L nodes encode a more complicated set of transitions between Ms in adjacent columns, while keeping the entire graph to a low degree.

(a) The right scanline in disparity space

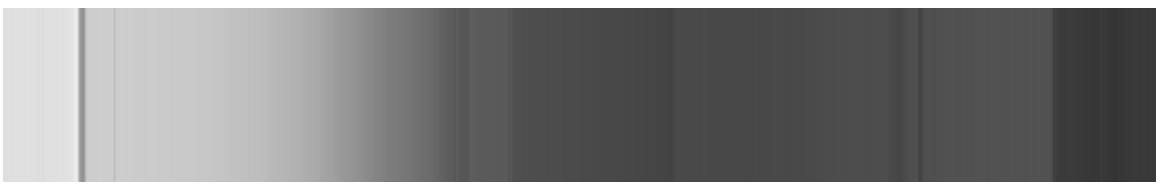

(b) The left scanline in disparity space

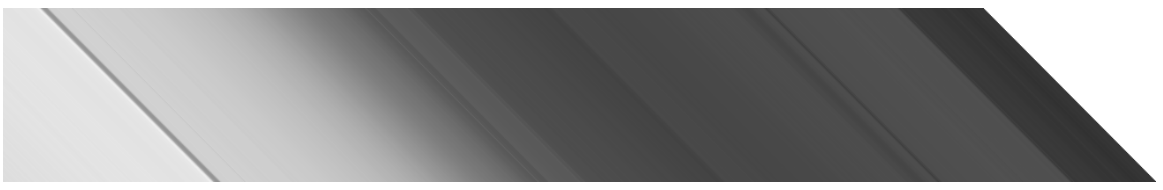

(c) The DSI combines these

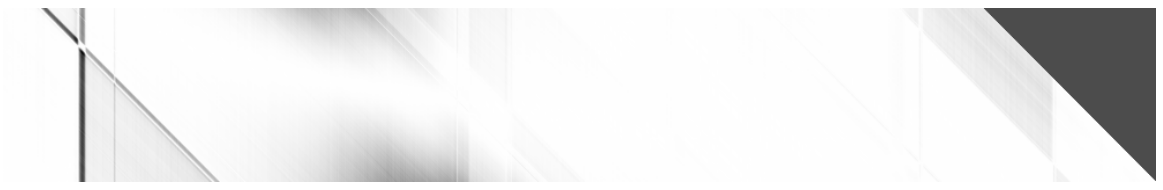

(d) Shortest Path overlayed

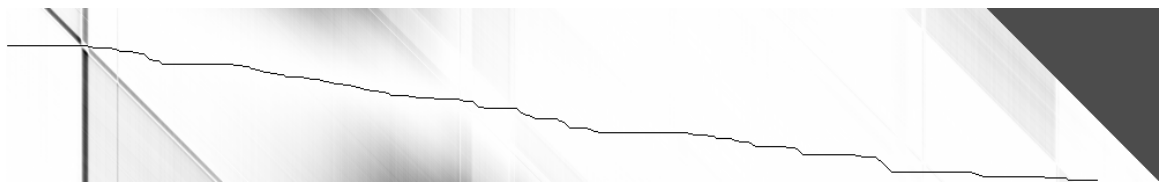

Figure 3: The DSI as an image: (a) The right scanline, (b) The left scanline, (c) The DSI, and (d) the shortest path.

structure. For each node in the DSI, we compute the incoming path costs in both the forward and backward directions. These can then be normalized (within $M$ and $R$ events) to produce a distribution over events for each pixel in the right image. This is valid because the $M$ and $R$ nodes are mutually exclusive and jointly exhaustive events for a given pixel in the right frame. While we have presented this algorithm from the perspective of the right camera, there is no loss of generality in doing so – values can be computed for the left camera in a similar manner.

In Section 2.2, we will present dynamic programming algorithms that use the sparseness of the DSI to compute the Viterbi path, marginal probability distributions, and path entropy for this model in $O(nd)$ time.

#### Creating the original DSI

To visualize the transition costs, Disparity Space Image can be viewed literally as an image. The top of the image represents high disparity values (objects that are closer to the camera), and the bottom of the image represents low disparity values. The image represents the cost function over the M nodes in the DSI. A pixel takes on the value of the transition cost into the M node at that (column, disparity) pair.

Every row in the DSI corresponds to a different disparity level at which pixels in the left image can match pixels in the right image. When the DSI is created from the perspective of the right camera, a scan line in the right image is projected into disparity space by simply repeating the image scanline $d$ times (Figure 3a). A left scan line is projected into disparity space by sliding it horizontally as the disparity level (vertical dimension in the DSI) changes (Figure 3b). The final DSI is then the difference between the these right and left images, with a cost function applied (Figure 3c). These scores become unnormalized log probabilities of transition into the M nodes in the graph view of the DSI. On an intuitive level, the most probable path through the DSI image is the light band (or dark, depending on your choice of scaling) visible where matchings are good (Figure 3d). This path is found by dynamic programming (Section 2.2).

#### Updating the DSI as laser results come in

At naturally occurring strong boundaries in the image pair (idealized as a two-tone vertically separated field in Figure 4), the DSI takes on a slanted X-shaped pattern, with the ideal matching corresponding to a path through the crux of the X. If the boundary is strong enough, this divides the shortest path problem into two independent problems. When we get a matching from our laser aim, we would like to get the same problem-splitting effect. We achieve this by filling nodes with impossibly high transition costs with the same X pattern, funneling all paths through the one matching point.

### 2.2 The DP Stereo Algorithm

Each node in the graph-DSI has *at most* three predecessor nodes and three successor nodes. Because of the sparse nature of the graph, we can calculate the shortest path through it in linear time using dynamic programming. With some slight modification to the dynamic programming algorithm, we can also calculate the marginal probability of any node

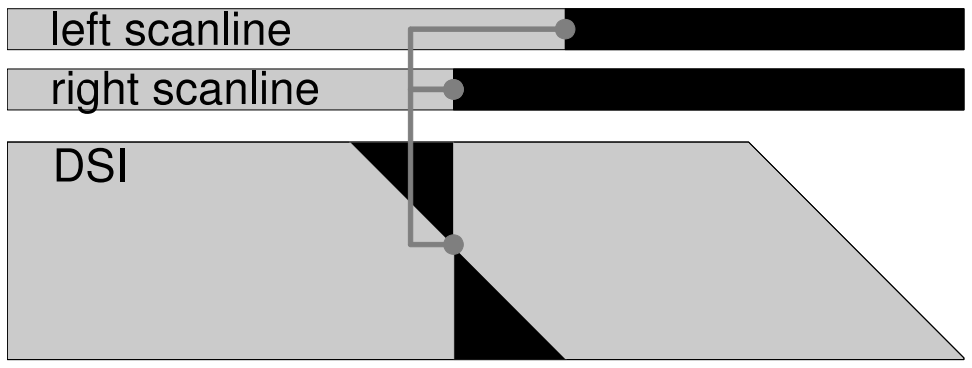

Figure 4: An X pattern in the DSI naturally generated at edges in the image. The effect is reproduced when the laser identifies a match. All nodes internal to the two triangles are blocked. Transitions to along the boundary are assigned very low probability. The $L$ nodes above the established $M$ node and $R$ nodes following the $M$ down the diagonal are exceptions, as discussed in detail in section 2.4.

(a)

| node $c$ | $\Gamma^-(c)$ | | |
|---|---|---|---|
| (i,j,M) | (i-1,j,M) | (i-1,j,L) | (i-1,j,R) |
| (i,j,R) | (i-1,j-1,M) | | (i-1,j-1,R) |
| (i,j,L) | (i,j+1,M) | (i,j+1,L) | |

(b)

| node $c$ | $\Gamma^+(c)$ | | |
|---|---|---|---|
| (i,j,M) | (i+1,j,M) | (i,j-1,L) | (i+1,j+1,R) |
| (i,j,R) | (i+1,j,M) | | (i+1,j+1,R) |
| (i,j,L) | (i+1,j,M) | (i,j-1,L) | |

Figure 5: Each node has at most three predecessor nodes (a) and three successor nodes (b).

and the path entropy through any node in linear time. These modifications require running the dynamic program both forwards and backwards over the graph. In the forward direction, the algorithm moves in the direction of the arcs in the DSI and computes a value for each node as function of its predecessor's values. In the backward direction, the algorithm moves against the direction of the arcs in the DSI and computes a value for each node as a function the node's successor values. The predecessor and successor sets, $\Gamma^-(c)$ and $\Gamma^+(c)$, respectively, are defined in Figure 2.2.

**Dynamic Programming to find the shortest path**

We begin by reconstructing the Bobick and Intille algorithm as a Viterbi path calculation through a graph. To find the best path, we iterate over the nodes in a column from bottom to top, moving left to right through the DSI. For each node, we consider the legal transitions into the node and select the lowest cost. We store this value as well as backward pointer indicating which predecessor gives us the value.

$$sp(c) = score(c) + \min_{b \in \Gamma^-(c)} sp(b)$$

At the end of the forward DP pass, we select the shortest path for all exit nodes and, using the backwards pointers, reconstruct the path which provides this lowest cost. If we view scores as negated log probabilities, the path with minimum score is equivalent to the path with highest probability, i.e., the Viterbi path.

**Modification to find marginals and path entropy**

The marginal probabilities over events in the DSI can be computed by an adaptation of the standard forward-backward procedure for HMMs:

$$p_f(c) = e^{-score(c)} * \sum_{b \in \Gamma^-(c)} p_f(b)$$

$$p_b(c) = e^{-score(c)} * \sum_{b \in \Gamma^+(c)} p_b(b)$$

$$p(c) \propto p_f(c) * p_b(c)$$

After running DP backwards and forwards, we normalize the probabilities over $M$ and $R$ for each pixel. To calculate the entropy of all paths running through a node we run the dynamic program:

$$h_f(c) = p_f(c) \sum_{b \in \Gamma^-(c)} (p_f(b) \log(p_f(c)) + h_f(b))$$

$$h_b(c) = p_b(c) \sum_{b \in \Gamma^+(c)} (p_b(b) \log(p_b(c)) + h_b(b))$$

$$h(c) = p_b(c) h_f(c) + p_f(c) h_b(c)$$

The cross-multiplication in each part stems from the identity $ab \log(ab) = a(b \log(b)) + b(a \log(a))$. To calculate the total path entropy of the system, we only need to run this in one direction, and take the sum over the end states.

### 2.3 Queries and Query Selection

In our framework, we begin by capturing a pair of reference images. A query corresponds to pointing a laser line generator at a specific column in one frame of the image. We assume that the laser is mounted on a pan/tilt mechanism centered directly above the nodal point of one of the camera lenses. Under this assumption, the laser will generate a nearly perfect vertical line in this camera's field of view. Note that the laser line will *not necessarily* be a vertical line in the other frame, but may appear as a sequence of line segments.

We isolate the laser line by subtracting the reference image from the images with the laser lines. Only the laser lines (and perhaps some noise) will remain. We identify the brightest point in each row of each image and this establishes one point of known correspondence in each row.

In practice this can be trickier than it sounds if there are specular surfaces, or conditions that lead to poor signal to noise ratio in the area hit by the laser.

Since there is a bijection between paths through the DSI and stereo matchings, the path entropy is a natural measure of our confusion about the best matching. We therefore select a query that maximizes the expected reduction in path entropy, the information gain (*IG*). In section 2.2, we present a modification to the shortest path algorithm that can calculate the total path entropy over the DSI in linear time; this dynamic program, run backwards and forwards, can also calculate the information gain of the available queries in linear time. In practice, however, the constant factor for this is large and we can save time by drawing upon the recent work of Anderson and Moore [1] for finding the path entropy minimizing query in HMMs. They note that, for a set of paths $\Pi$, query $Q_i$, and state $S_i$, the following are equivalent:

$$\begin{aligned} IG(Q_i) &= H(\Pi) - H(\Pi|Q_i) \\ &= H(Q_i) - H(Q_i|\Pi) \\ &= H(Q_i) - H(Q_i|S_i) \\ &= H(S_i) - H(S_i|Q_i), \end{aligned}$$

where the transition from the second to third step follows from the Markov property. $H(S_i)$ corresponds to the entropy in the marginal distribution over the $M$ and $R$ nodes, and $H(S_i|Q_i)$ is the expected conditional entropy after the observation. If our query returns a matching, the entropy drops to 0 for that column. If our query returns an occlusion, we can't say which $R$ node we are in, but we can calculate the entropy over the renormalized distribution over the $R$ nodes, as we can be sure we are not in an $M$ node. It is possible for our entropy to increase significantly if we are expecting a matching but are presented an occlusion.

### 2.4 Theory vs. Reality

Our approach was initially developed on artificial images with artificially generated laser lines based on ground truth. In applying our method to the real world, we had to make a few modifications to ensure that the algorithm would continue to function.

#### Accepting 'Catch up' Occlusions

When a flat surface appears at an angle to the image plane, it will take up more room in one of the images compared to the other. In the DSI, matchings can only occur along a constant disparity level, and occlusion steps are required to change disparity levels. Angled surfaces are represented as a (finely grained, if the algorithm is close to correct) series of frontoparallel surfaces interspersed with catch up occlusions.

Ideally, when a laser line indicates that two points match in the stereo pair, we would like the shortest path to include the corresponding M node. However, with angled surfaces, there might not be a one-to-one matching between pixels.

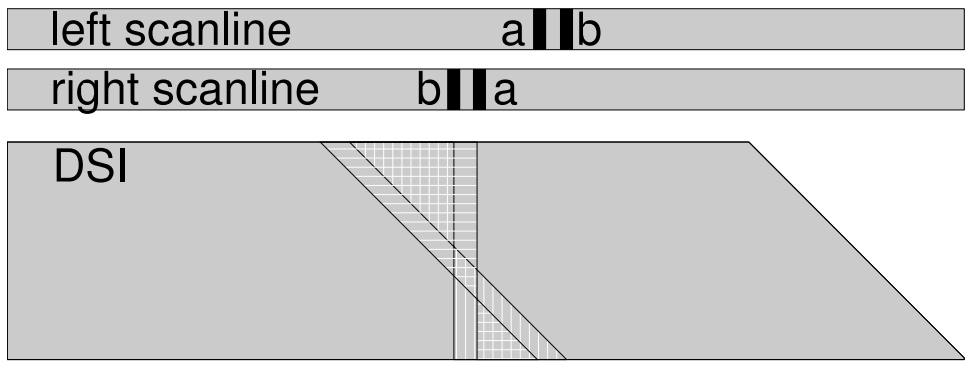

Figure 7: When two matching sets of pixels reverse their order between two scanlines, they violate the ordering constraint. In the DSI, this manifests as two points within each others' dead zone (see figure 4), meaning there is no valid path between them.

In particular, when two laser aims are close to each other, their results might call for some DSI-impossible paths such as the one shown in figure 6.

To solve this problem, we make a small modification to the laser updates to the DSI: All nodes along the borders of the X-shaped region remain possible, though unlikely. The M node identified by the laser is given a zero cost (high probability of match), the nodes internal to the X are given an insurmountable cost (probability near 0), and all the other nodes in the same column and diagonal as the M are given a large, but not insurmountable cost. This has the effect of permitting local violations of the uniqueness constraint if (and only if) they are the only explanation consistent with the laser data.

#### Detecting Violations of the Ordering Constraint

The ordering constraint states that if two pixels $a_r$ and $b_r$ in the right image match pixels $a_l$ and $b_l$ in the left image, then those pixels must occur in the same order in each image. In the real world, this can be violated when, for example, there is a thin object not connected to objects behind it near the camera. A graceful failure mode for this case is to match the close, thin object and use occlusions on either side to avoid violating the ordering constraint. However, the laser can prevent this by explicitly establishing matches that violate the ordering constraint. Naively entering these matches into the DSI can have the effect of rendering all paths through the DSI impossible.

To detect this case, we note that, in the slanted X we create in the DSI when we establish a matching, the forbidden zones we establish correspond to violations of the ordering constraint (see Figure 7). As new matchings come in, we first check if they are within a forbidden zone from an earlier query. If we were willing to accept the increase in memory or computation time, the ideal solution to this problem would be to accept the closer object as the correct one, reverting to or recreating the parts of the DSI changed by the earlier query result, but in our current implementation, priority is given to the earlier update to the DSI.

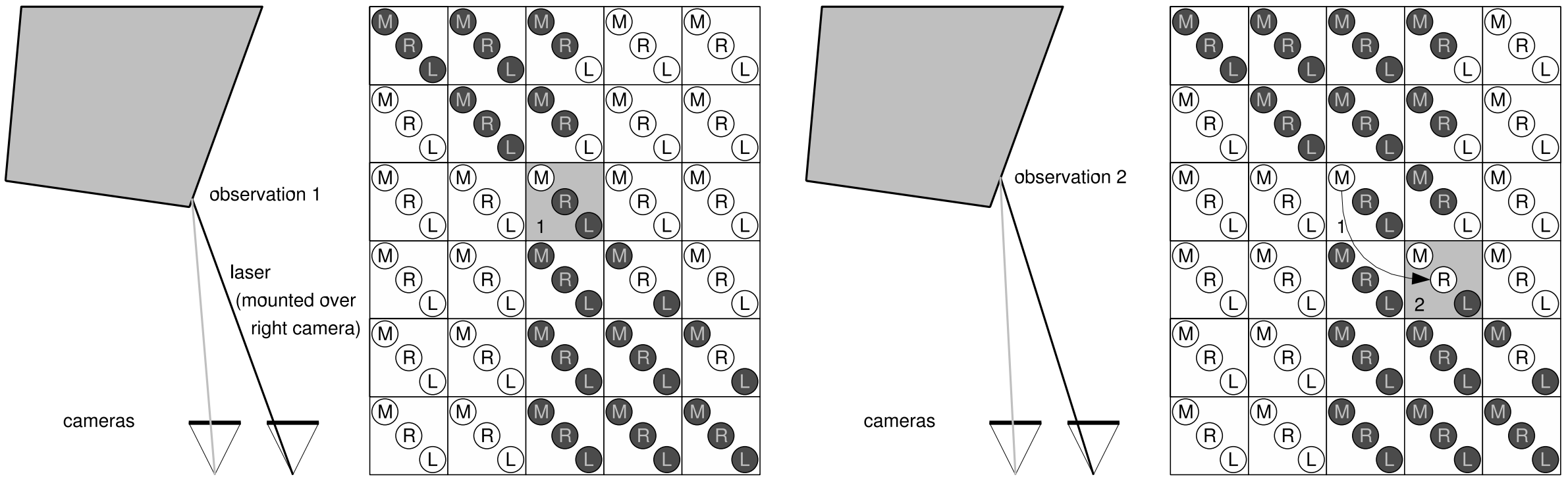

(a) observation 1 (b) observation 2

Figure 6: As the laser moves along a sheer surface, it can occasionally match two pixels in one image to one pixel in the other. In this case, the observations in (a) and (b) lead to two adjacent pixels in the right image matching to one pixel in the left image. These two $M$ nodes are not joined by any valid path, and so the algorithm must accept paths through the $R$ node at the same disparity level as the second matching.

# 3 Synthetic Experiments

In this section we present results using stereo pairs for which ground truth data are available. We simulated a laser line projected onto these scenes. Please note that the subtle differences in grayscale corresponding to depth changes are *much* easier to see on screen than in print.

## 3.1 Rendered Images

In our first set of synthetic experiments, we used a rendered pair of $1256 \times 810$ images (Figure 8 a,b) that were intentionally created with low texture information. Ground truth data were available from the rendering program and these were used to simulate laser lines. Figure 8c shows the ground truth disparity map for this image pair. Figure 9a shows the initial disparity map from the stereo algorithm. A disparity map shows pixel disparities as luminance. Higher disparities correspond to closer depths and higher luminance in the disparity map. Notice the the floor close to the camera is totally wrong due to the lack of texture information. Figure 9b shows the disparity map after 9 laser aims and Figure 9c shows the entropy map after 9 laser aims. The entropy map is like the disparity map, but shows areas of high entropy (in the marginal distribution) with higher luminance. The dark vertical bands show laser aim points, in which the entropy has been forced to zero. The algorithm has chosen the areas of low texture that will have the greatest expected reduction in total path entropy.

To help quantify the effect of our laser aiming strategy, we provide two graphs. In Figure 10a, we show the total path entropy through the DSI as a function of the number of laser aims and in Figure 10b, we show the number of pixels with disparity errors greater than 1. In both cases we compare against an average of 10 random laser aims. Random aims can, initially, do well in this image because nearly anything helps resolve the large, ambiguous floor area close to the camera. However, our strategy of maximizing expected information gain establishes a growing lead after the first few aims.

## 3.2 Benchmark Images

We performed experiments on several of the benchmark images from the Middlebury stereo vision suite [7]. We briefly present some results for several of these images in Figure 11. Maximizing information gain generally outperforms random aims for entropy reduction, although the benefit is not always large. For pixel error (the number of pixels off by more than 1 disparity value), there is no consistent advantage to maximizing information gain - at least for a horizon of 9 laser aims. Neither of these results are unexpected. These benchmark images are well textured to start with, so adding additional texture, no matter how well planned, will have limited benefit. Since our algorithm directly optimizes information gain, we would expect it to perform best for entropy reduction, which doesn't necessarily imply short term improvements in pixel error. For example, the laser aims could serve to confirm choices that were (fortuitously) correct in the Viterbi matching without having much impact on the pixel error.

# 4 Physical Implementation and Experiments

To test our algorithm on a realistic scenario more similar to what a robot would actually encounter, we built a prototype system (Figure 12a) and moved it into the hallway of our department (Figure 12b). The prototype system consisted of two consumer digital SLR cameras attached to a tripod, a computer controlled pan/tilt head connected to a separate tripod, and a consumer green laser pointer with an inexpen-

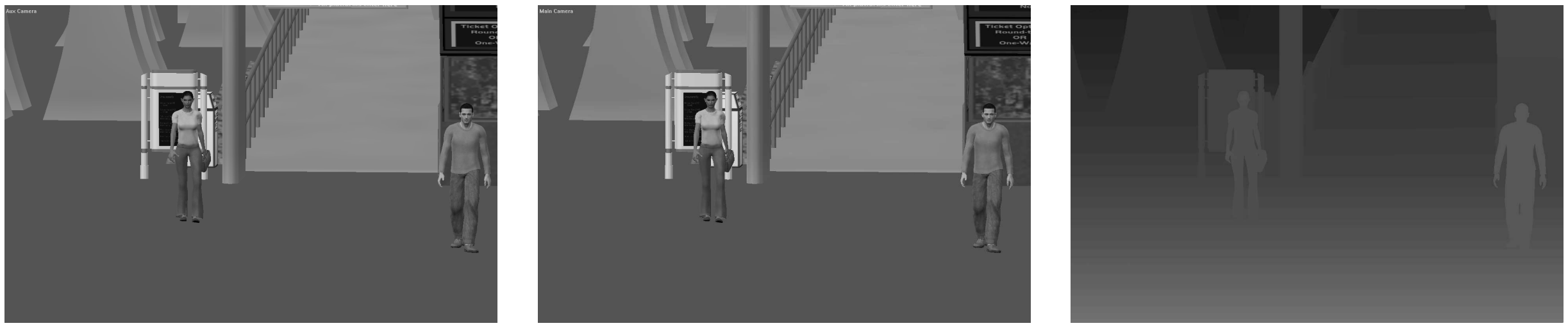

a b c

Figure 8: (a) A rendered scene with low texture (left camera view), (b) right camera view, (c) Ground truth disparities. Lighter is closer.

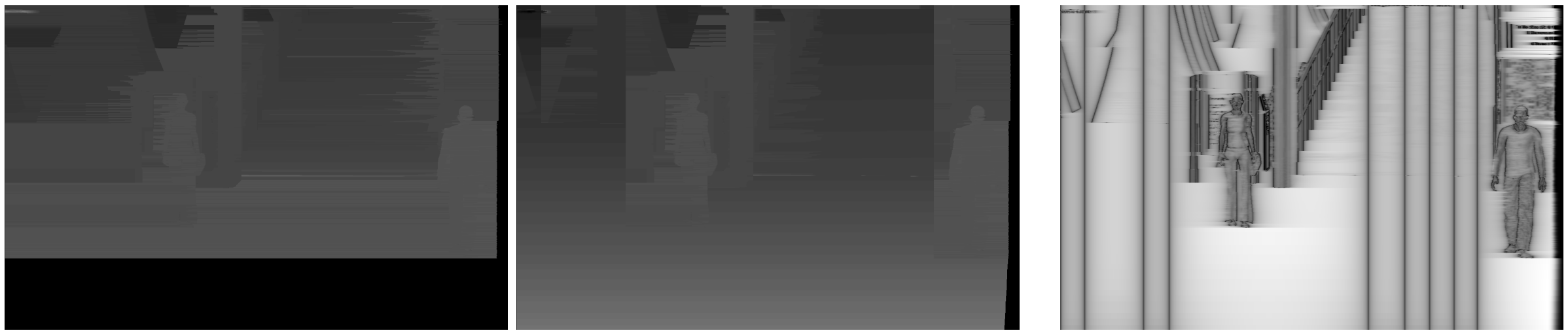

Figure 9: (a) The initial disparity map, (b) The disparity map after 9 laser aims, (c) The entropy map after 9 laser aims. Dark vertical bands show aim points for the laser.

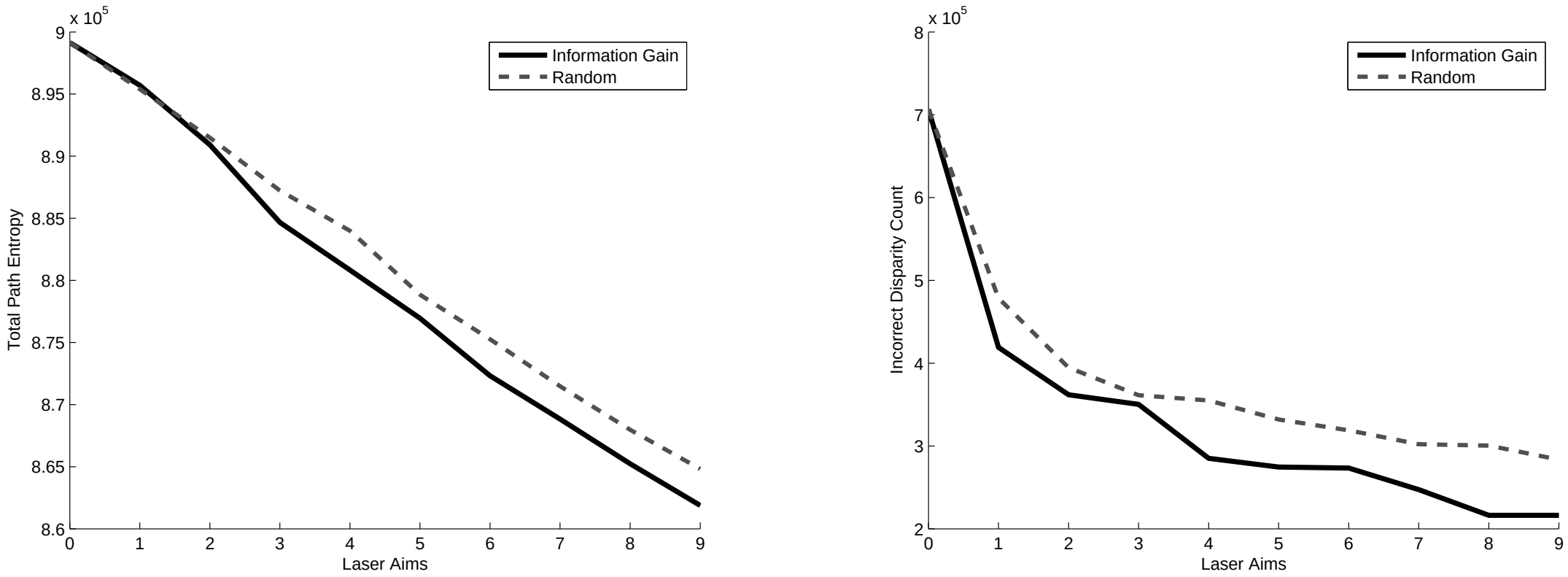

(a) (b)

Figure 10: Rendered images: (a) Total path entropy as a function of laser aims, information gain vs. random aims, (b) Pixels with greater than 1 error in disparity, information gain vs. random aims

| | Avg. Path Entropy Reduction | | Avg. Pixel Error Reduction | |
|---|---|---|---|---|
| Dataset | Information Gain | Random | Information Gain | Random |
| Tsukuba | 2220 | 1886 | 882 | 893 |
| Venus | 2860 | 2461 | 1344 | 1272 |
| Sawtooth | 1088 | 994 | 60 | 217 |
| Cones | 6600 | 3255 | 219 | 209 |

Figure 11: Summary of performance on benchmark stereo pairs.

sive beam spreading lens attached to the front. The entire apparatus was connected to a laptop. The apparatus may appear a bit bulky but this is largely due to some choices that were made to permit faster prototyping. A real system on a robot could use smaller cameras and a less advanced pan/tilt unit since high pan/tilt accuracy is not critical for our application.

The cameras were carefully calibrated before the experiments, but the only calibration of the laser was some hand adjustment to ensure that it looked approximately vertical in the right frame. We performed laser detection with image subtraction and some simple heuristics.

We generated a "ground truth" disparity map by producing approximately 200 laser aims and using stereo vision to fill in the gaps between the laser aims. The data collection took approximately 90 minutes of real time and the results are shown in Figure 12c. The disparity map matches our personal knowledge of the scene quite well. The difference between this and the initial disparity map shown in Figure 13a is striking, due to the textureless walls. The initial disparity map interprets the slanted wall as a collection of panels parallel to the image plane, separated by changes in luminance. This is a common artifact of stereo algorithms.

We applied our entropy minimization strategy to the data set collected in our own hallway. The images with 200 laser aims were treated as ground truth and queries were satisfied by returning the closest of the 200 laser aims to the request made by the algorithm. The results of these runs, compared to random laser aims, and evenly spaced laser aims are shown in Figure 15. As expected, entropy minimization significantly outperforms the alternatives when the performance criterion is the reduction in path entropy. In the short term, entropy minimization does not seem to perform well at pixel error reduction, but it appears that the cumulative effect of entropy reduction pays off with more laser aims, as seen on the right hand side of Figure 15b. The corresponding disparity maps are shown in Figure 13 b,c. While the disparity map still isn't perfect, the effect of 9 laser aims is a substantial improvement. Figure 14 a-c shows the entropy maps before and after laser aims. The aim points chosen by the algorithm correspond well with weakly textured, highly ambiguous areas of the image.

## 5 Related Work

It is not uncommon in the stereo vision literature for matching costs to be interpreted as probabilities [2, 5, 4, 9]. However, such interpretations are typically seen primarily as justifications for various optimization techniques. The use of structured light sources from a projector is also a fairly well established technique [8].

The observations made by Anderson and Moore [1] are most relevant to our information gain computation, but the work of Guestrin and Kraus [6], which considers optimal nonmyopic information gathering actions is also highly relevant. Unfortunately, the algorithms presented in that work, while polynomial for structures like ours, are still too slow for real time use. For a large image, anything more than $O(nd)$, even $O(nd^2)$, is impractical since $d$ can be quite large (hundreds of pixels).

Our work can be viewed as one of the first that uses the probabilistic interpretation of stereo matching costs for some purpose other than match cost optimization. We offer the first probabilistic interpretation of the Bobick and Intille algorithm, generalize this algorithm to compute marginal probabilities and entropies efficiently, and apply insights from graphical models to compute the information gain efficiently.

## 6 Future Work

A most important practical direction for future work is the full deployment of this technique on a robot. This would most likely entail the use of some compact cameras and a simpler pan tilt mechanism to reduce overall bulk. Once this is accomplished, we would like to integrate the new sensor into a vision based mapping algorithm.

For the sensor itself, there are several interesting directions for future work. First, our algorithm uses a laser line, but a laser line may not be practical in all cases. Due to eye safety concerns, it may not always be possible to use a laser powerful enough to generate a line that is spans the entire vertical field of view and is visible in bright light. A natural solution is to use a beam spreader with narrower dispersion and to use the tilt feature of a pan/tilt head to choose how to aim the laser line segment vertically. This is a fairly straightforward generalization of our approach which can be achieved quite efficiently.

From the algorithmic standpoint, another practical consideration is that total path entropy may not be the best criterion to optimize. Initially, it was a choice of (computational) convenience. We are investigating if other criteria can be computed as efficiently in some cases. We are also interested in the case where some regions of the image are identified as more important than others and the optimality criterion is weighted accordingly.

Another direction for exploration would be the use of a more sophisticated stereo model. A natural fit would be the belief propagation approach [9]. However, it is not guaranteed that using a more sophisticated stereo model would be worth the challenges involved, since the algorithm itself would be significantly slower and it would be difficult or impossible to compute the information gain efficiently. While a more sophisticated stereo model could in turn provide better probabilities that could provide better guidance for a laser aiming strategy, it could not fully resolve the fun-damental ambiguity posed by textureless surfaces without making some strong, additional assumptions. While this direction is worth exploring, it could turn out that using a simpler stereo algorithm is more efficient overall.

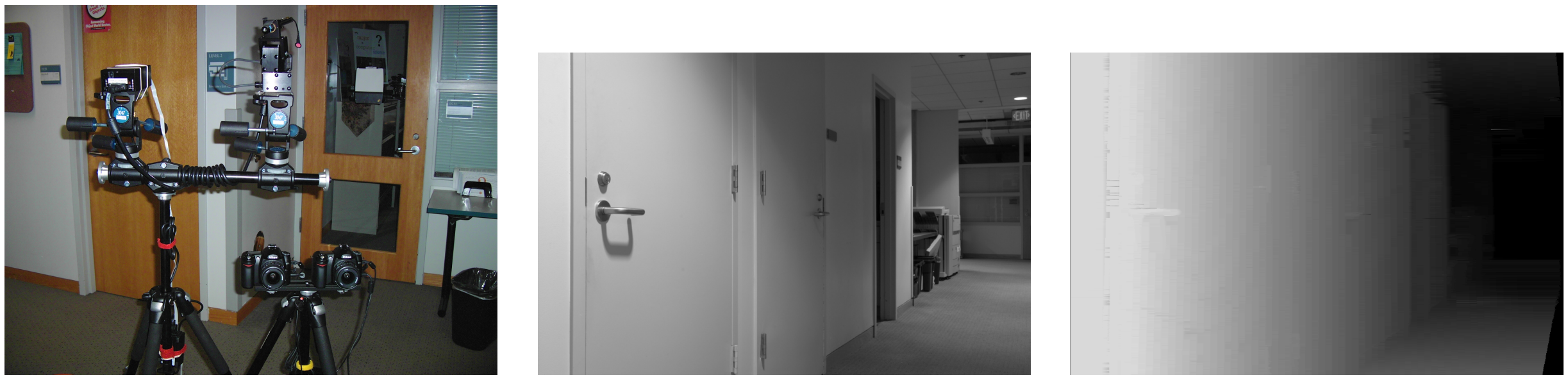

(a) (b) (c)

Figure 12: (a) Our prototype camera/laser apparatus, (b) Our hallway, (c) Disparity map after 200 laser aims.

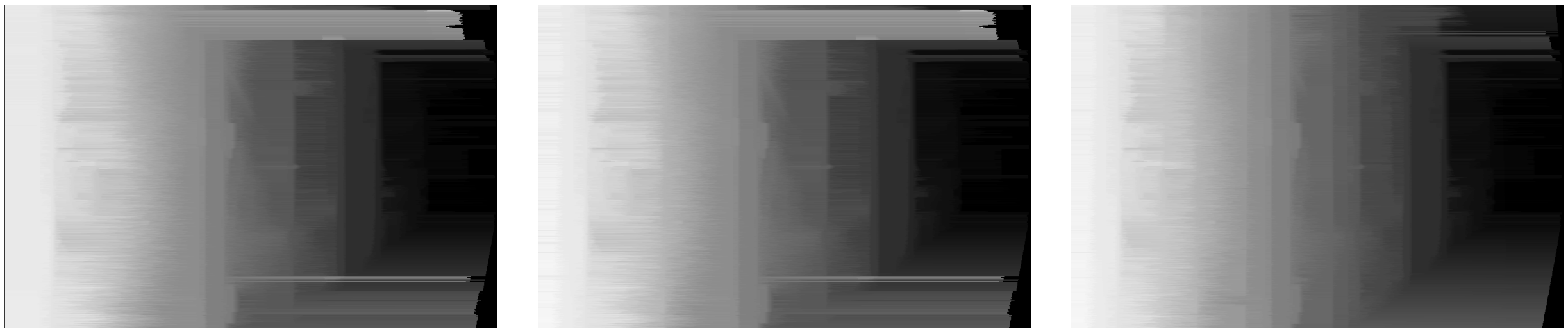

(a) (b) (c)

Figure 13: Hallway: (a) The initial disparity map, (b) The disparity map after 2 laser aims, (c) The disparity map after 9 laser aims.

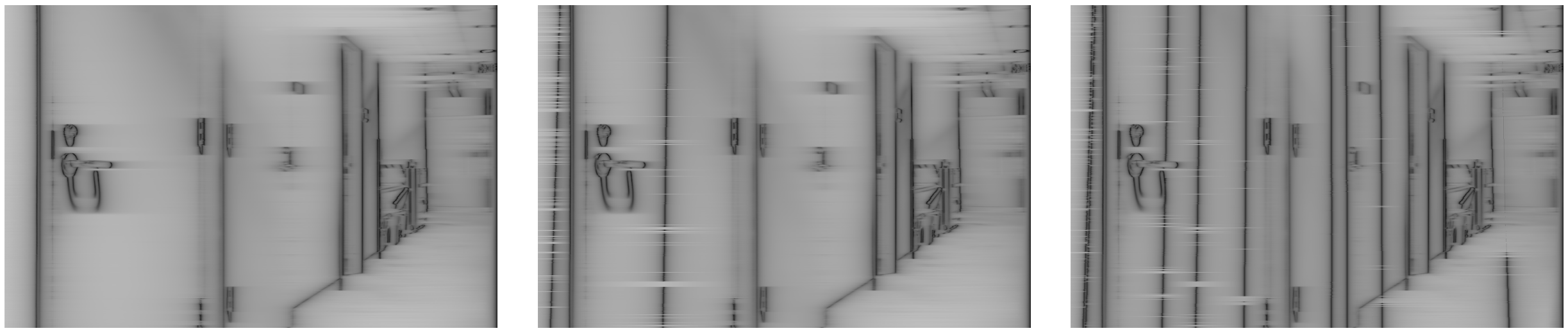

(a) (b) (c)

Figure 14: Hallway: (a) The initial entropy map, (b) The entropy map after 2 laser aims, (c) The entropy map after 9 laser aims. Gaps in the vertical lines from the laser can be due to occlusions or areas where the laser could not be detected with high confidence.

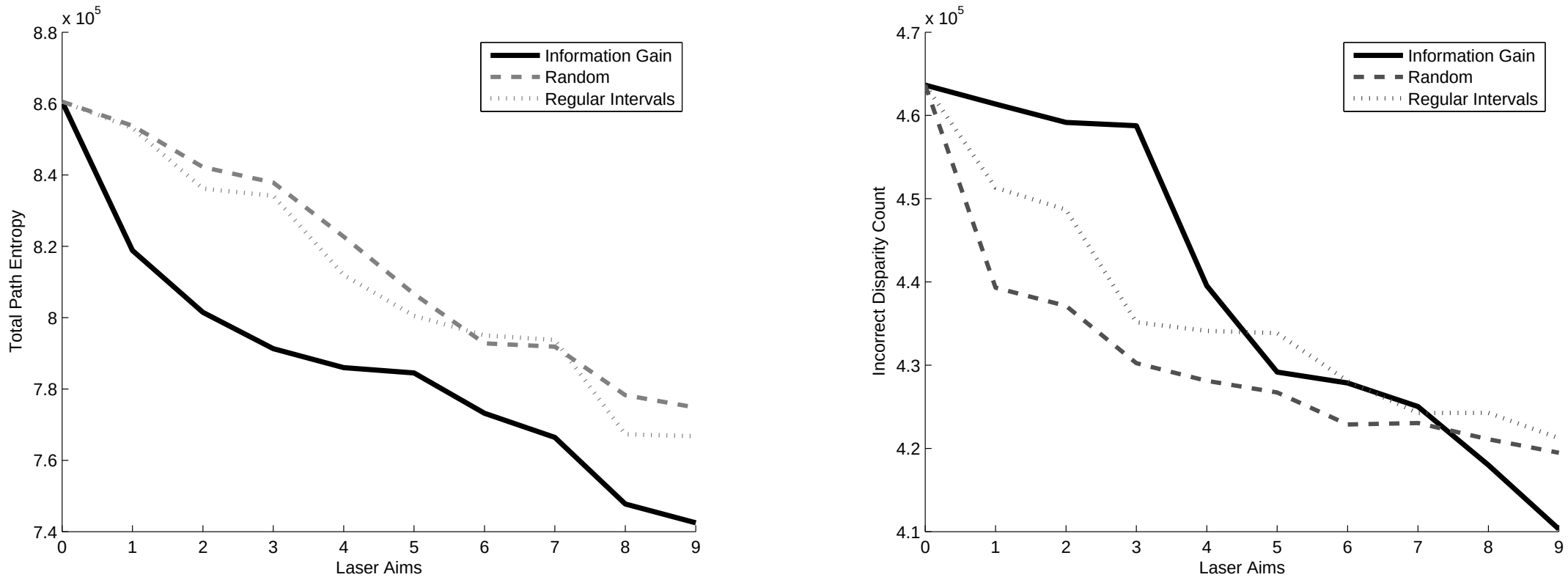

(a) (b)

Figure 15: Hallway: (a) Total path entropy vs. number of laser aims, (b) Pixels off by more than one disparity value vs. laser aims

Finally, a less glamorous but no less important area for further study is the choice of parameters for the algorithm. This is an issue for stereo algorithms in general and it is more of a concern for our approach. In addition to the occlusion penalties, we must choose a scaling factor when converting scores to potentials. The choice of scaling factor can make the distribution over paths more or less peaked and can alter the behavior of the algorithm. Different size images appear to require different parameters and a principled method for determining these parameters would be a significant improvement over our ad hoc search for good values.

## 7 Conclusions

We have addressed the challenge of active stereo vision using an entropy minimization approach. By adopting a probabilistic interpretation of an existing $O(nd)$ stereo algorithm and adapting this algorithm to compute probabilities and entropies, we have devised an approach to selecting the action with the greatest information gain that is, asymptotically, no more expensive than the core stereo algorithm. This is critical for the stereo problem because even a quadratic cost in the maximum disparity can be extremely large for high resolution images.

Our approach to this problem leverages a probabilistic interpretation of the stereo problem and employs insights gained from recent work probabilistic reasoning for sensor management.

We have implemented this algorithm and shown that it makes good choices of laser aim points in simulation of a hybrid stereo vision/laser device. We have also built this device and demonstrated that it can be used to produce high resolution disparity maps of real scenes. We are actively pursuing this approach as a practical alternative to prohibitively expensive and bulky three dimensional laser range finders.

## Acknowledgment

We are grateful to Carlo Tomasi for many helpful suggestions. The rendered image used in our synthetic experiments was provided by IAI. This work was supported by NSF IIS award 0209088, SAIC, and the Sloan Foundation.